\documentclass{article}
\usepackage{spconf,amsmath,graphicx}
\usepackage{subfigure}
\usepackage{amssymb}
\usepackage{times}
\usepackage{epsfig}
\usepackage{amsmath}
\usepackage{amssymb}
\usepackage{multirow}
\usepackage{nidanfloat}
\usepackage{diagbox}
\usepackage{array}
\usepackage{booktabs}
\usepackage{caption}
\usepackage[dvipsnames]{xcolor}
\newcommand{\cjj}[1]{\textcolor{black}{#1}}
\newcommand{\jy}[1]{\textcolor{black}{#1}}

\title{JNDMix: JND-Based Data Augmentation for No-reference Image Quality Assessment}
\name{Jiamu Sheng$^1$, Jiayuan Fan$^*$$^1$\thanks{$^*$Corresponding author.}, Peng Ye$^2$, Jianjian Cao$^2$\thanks{This work was supported in part by the National Natural Science Foundation of China under Grant 62101137 and Grant 62071127, and in part by the Zhejiang Lab Project under Grant 2021KH0AB05.}}
\address{$^1$Academy for Engineering and Technology, Fudan University, Shanghai, China\\
$^2$School of Information Science and Technology, Fudan University, Shanghai, China}
\begin{document}
%
\maketitle
\begin{abstract}
Despite substantial progress in no-reference image quality assessment (NR-IQA), \jy{previous training} models often suffer from over-fitting due to the limited scale of \jy{used} datasets, resulting in model performance bottlenecks. To tackle this challenge, we explore the potential of leveraging data augmentation to improve data efficiency and enhance model robustness. However, most existing data augmentation methods incur a serious issue, namely that it alters the image quality \jy{and leads to training images mismatching} with their original labels. Additionally, although only a few data augmentation methods are available for NR-IQA task, their ability to enrich dataset diversity is still insufficient. \jy{To address these issues, we propose a effective and} general data augmentation based on just noticeable difference (JND) noise mixing for NR-IQA task, named JNDMix. In detail, we randomly inject the JND noise, imperceptible to the human visual system (HVS), into the training image without any adjustment to its label. Extensive experiments demonstrate that JNDMix significantly improves the performance and data efficiency of various state-of-the-art NR-IQA models and \jy{the commonly used} baseline models, as well as the generalization ability. More importantly, JNDMix \jy{facilitates MANIQA to} achieve the state-of-the-art performance on LIVEC and KonIQ-10k.

\end{abstract}
\section{Introduction}
\label{sec:intro}
No-reference image quality assessment (NR-IQA) is a significant task to automatically predict image quality scores without using reference image information. With the development of deep learning techniques, a large number of learning-based NR-IQA methods achieve \jy{good} performance. Using pre-trained models \jy{based} on a large-scale dataset like ImageNet \cite{deng2009imagenet} can help NR-IQA models to extract relevant features for distorted images with the limited scale of \jy{used} datasets. Therefore, many \jy{recent works~\cite{zhang2018blind,su2020blindly,yang2022maniqa,ye2022beta,ye2022efficient}} utilize ImageNet pre-trained backbone to achieve \jy{the} state-of-the-art performance on the NR-IQA benchmarks.  
\begin{figure}
\centering
\includegraphics[width=0.41\textwidth]{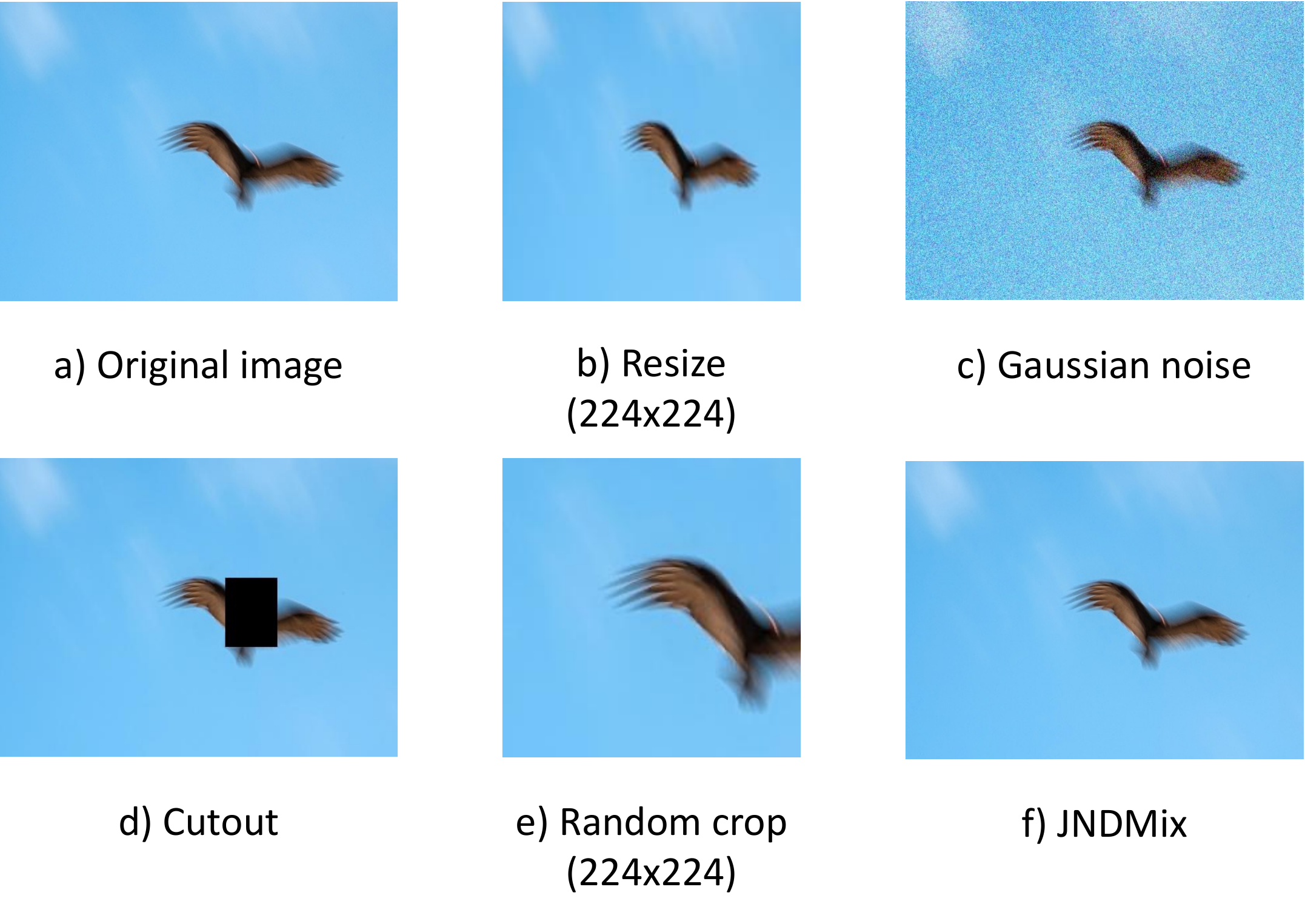}
\vspace{-3mm}
\caption{\jy{An original image is processed by five different data augmentation methods. Visually, the other four data augmentation methods cause highly noticeable image quality degradation, except JNDMix.}}
\vspace{-4mm}
\label{fig:aug}
\end{figure}

\par
A key challenge of learning-based NR-IQA methods is the lack of sufficiently large labeled datasets, as manually annotating image quality labels is expensive and time-consuming. As a result, the learning-based NR-IQA methods usually suffer from over-fitting when training deep models \cite{zhu2020metaiqa}. To some extent, using pre-trained models can alleviate this issue, but this metric is not specifically designed for NR-IQA task, and the model performance bottleneck still exists. Therefore, it is necessary to design \jy{a novel approach} to improve the data efficiency of the NR-IQA model.

Data augmentation is a practical and simple way to improve generalizability and data efficiency of models, which can effectively alleviate the above problems on NR-IQA task. \jy{So far, data augmentation methods are widely used in various vision tasks, that} include geometric transformations (e.g., random cropping \cite{yeh2022saga}, horizontal flipping \cite{mclaughlin2015data}, cutout \cite{devries2017cutout}), color space transformations (e.g., color jittering \cite{szegedy2015going}), 
mixing images (e.g., mixup \cite{zhang2018mixup}, CutMix \cite{yun2019cutmix}), \jy{resizing \cite{girshick2018detectron},  noise injection \cite{moreno2018forward}}, etc. 
Unfortunately, most of data augmentation methods mentioned above, \jy{are mainly developed for specific vision tasks and not available for the NR-IQA task, since the highly noticeable image quality degradation is caused \cite{ke2021musiq}. As shown in Fig.~\ref{fig:aug}, noise injection with Gaussian noise directly introduces distortions to the image; resizing, cutout and random cropping can impact image composition, thus altering the image quality.} 
Although a few augmentation methods such as horizontal flipping and random cropping are widely used in NR-IQA task \cite{yan2018two}, they are not specifically designed for NR-IQA task and are still limited in their effectiveness in enriching dataset diversity. More seriously, random cropping leads to the loss of image information and introduces geometric deformation, resulting in generated patches mismatching with their labels, while existing models using random cropping ignore the above problem and still use the original labels to train. Hence, a general and effective data augmentation method is urgently needed for the NR-IQA task.

\jy{Intuitively, it is reasonable that the noise injection augmentation without noticeable image quality degradation is suitable for NR-IQA task. As the characteristic of the human visual system (HVS), just noticeable difference (JND)} is defined as the minimum visual content changes that human can perceive \cite{hall1977nonlinear}. Namely, for each pixel, JND is a threshold, and the HVS will not perceive any changes under it. Hence, we believe the change under JND as a kind of noise, named JND noise, which can be used in noise injection augmentation. In NR-IQA datasets, the labels are the quality of images subjectively scored by the human. Due to the imperceptibility of JND noise, the image injecting JND noise will have the same subjective quality score as the original image.

In this paper, we are the first to propose a general data augmentation method based on JND noise mixing (JNDMix) specifically for NR-IQA task. In detail, we randomly inject JND noise to the training image, and it will not alter the image quality visually so that the generated image can remain the original label. The seminal proposal of JNDMix tackles the long-standing problem with existing data augmentation methods available for NR-IQA task, in particular, further enriching the diversity of training images. Hence, JNDMix can help NR-IQA models effectively to improve model performance while enhancing data efficiency over previous state-of-the-art methods. We conduct extensive experiments on two IQA benchmarks, LIVEC and KonIQ-10k. And the results demonstrate that JNDMix bring the significant improvements in performance and data efficiency of various state-of-the-art NR-IQA models and baseline models. More importantly, our proposed JNDMix helps MANIQA to achieve the state-of-the-art performance on LIVEC and KonIQ-10k. The cross-dataset evaluation results also verify the great generalization ability of our proposed method for NR-IQA models. 
Besides, our approach is general and can be flexibly added to any NR-IQA method.

\section{Method}
\label{sec:method}

\begin{figure}
\centering
\vspace{-4mm}
\includegraphics[width=0.5\textwidth]{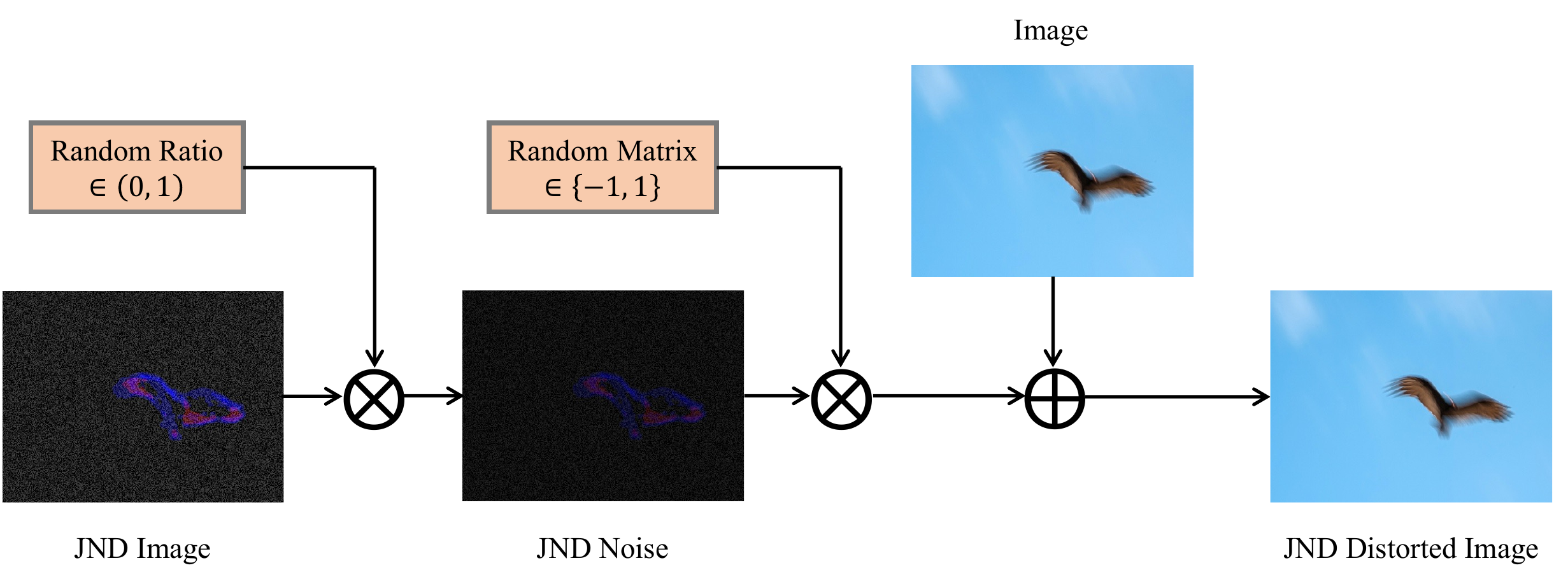}
\caption{Overview of JNDMix data augmentation workflow.}
\vspace{-3mm}
\label{fig:env}
\end{figure}

In this section, we will present the details of our proposed general data augmentation method, depicted in \jy{Fig.}~\ref{fig:env}. 
\subsection{Random JND Noise Injection Procedure}
Let $x \in \mathbb{R}^{W \times H \times C}$ and $y$ denote an original training image and its label, respectively. The goal of JNDMix is to generate a new training image $\tilde{x}$ and its associated label $\tilde{y}$. The original training image $x$ and the associated generated JND image $x_{\text {jnd}}$ are formed into an image pair to jointly generate a new training image. Firstly, each pixel of the JND image $x_{\text {jnd}}$ is multiplied by the random ratio $\lambda$ to generate a random JND noise $x_{\text {jnd}}$ where each pixel is less than the JND threshold. Then we inject our random JND noise $x_{\text {jnd}}$ into the associated training image. For each pixel, the injection noise can either add the noise to the original pixel value or subtract the noise, so we have to multiply a random matrix $r \in \mathbb{R}^{W \times H \times C}$ that is random on each pixel, with only $1$ and $-1$. Finally, the \cjj{novel} training image, named JND distorted image, is generated after the above process. This random JND noise injection procedure can be formulated as follows:
\begin{equation}
x_{\text {noise}}=\lambda x_{\text {jnd}}
\end{equation}
\begin{equation}
\tilde{x}=x \oplus(r \cdot x_{\text {noise}})
\end{equation}
where $\oplus$ is element-wise sum. $(\cdot)$ is element-wise product. Random ratio $\lambda$ is sampled from the uniform distribution $(0, 1)$. 
\subsection{Model Training}
In the training phase, the generated JND distorted image $\tilde{x}$ replaces the original training image $x$ in the model, but we do not need to adjust the associated quality score label $y$. The image quality scores in the dataset are obtained by subjective human scoring, while JND noise can not be perceived by humans theoretically, so the subjective scores of JND distorted image will not change. Thus, the following equation is obtained:
\begin{equation}
\tilde{y}=y
\end{equation}
Then, the new training sample $(\tilde{x}, \tilde{y})$ generated by JNDMix is used to train the model with its original training settings and loss function. In each training iteration, each JNDMix-ed sample $(\tilde{x}, \tilde{y})$ is generated by multiplying the JND image by a resampled random ratio to enrich the training sample's diversity, and diverse training samples reduce over-fitting. 
Our method is very general and concise, so it can be flexibly added to any NR-IQA method.

\section{Experiments}
\label{sec:experiments}

\begin{table}[t]
\footnotesize
\centering
\vspace{-2mm}
\caption{Model performance improvements with JNDMix across different \textbf{state-of-the-art models}.}
\vspace{-2mm}
\begin{tabular}{lrrrr}
  \hline\specialrule{0em}{1pt}{1pt}
  \multicolumn{1}{c}{\multirow{2}{*}{Method}} & \multicolumn{2}{c}{LIVEC} & \multicolumn{2}{c}{KonIQ-10k} \\ \cline{2-5}\specialrule{0em}{1pt}{1pt}
  \multicolumn{1}{c}{} & SRCC & PLCC & SRCC & PLCC \\ 
  \hline
  DBCNN (100$\%$) & 0.851 & 0.869 & 0.906 & 0.923 \\
  w/ JNDMix  & \textbf{0.854}  & \textbf{0.876} & \textbf{0.915} & \textbf{0.930} \\
      & \textcolor{blue}{\textbf{+0.003}} & \textcolor{blue}{\textbf{+0.007}} & \textcolor{blue}{\textbf{+0.009}} & \textcolor{blue}{\textbf{+0.007}}\\
  \hline
  HyperIQA (100$\%$) & 0.859 & 0.882 & 0.906 & 0.917 \\
  w/ JNDMix  & \textbf{0.870}  & \textbf{0.883} & \textbf{0.911} & \textbf{0.924} \\
      & \textcolor{blue}{\textbf{+0.011}} & \textcolor{blue}{\textbf{+0.001}} & \textcolor{blue}{\textbf{+0.005}} & \textcolor{blue}{\textbf{+0.007}}\\
  \hline
  MANIQA (100$\%$) & 0.886 & 0.908 & 0.923 & 0.941 \\
  w/ JNDMix  & \textbf{0.897} & \textbf{0.918} & \textbf{0.927} & \textbf{0.943} \\
      & \textcolor{blue}{\textbf{+0.011}} & \textcolor{blue}{\textbf{+0.010}} & \textcolor{blue}{\textbf{+0.004}} & \textcolor{blue}{\textbf{+0.002}}\\
  \hline
\end{tabular}
\label{tab:sota}
\end{table}
\begin{table}[t]
\footnotesize
\centering
\vspace{-1mm}
\caption{Model performance improvements with JNDMix across different \textbf{baseline models}.}
\vspace{-2mm}
\begin{tabular}{lrrrr}
  \hline\specialrule{0em}{1pt}{1pt}
  \multicolumn{1}{c}{\multirow{2}{*}{Method}} & \multicolumn{2}{c}{LIVEC} & \multicolumn{2}{c}{KonIQ-10k} \\ \cline{2-5}\specialrule{0em}{1pt}{1pt}
  \multicolumn{1}{c}{} & SRCC & PLCC & SRCC & PLCC \\ 
  \hline
  Resnet34 (100$\%$) & 0.824 & 0.849 & 0.905 & 0.920 \\
  w/ JNDMix  & \textbf{0.842}  & \textbf{0.868} & \textbf{0.910} & \textbf{0.928} \\
      & \textcolor{blue}{\textbf{+0.018}} & \textcolor{blue}{\textbf{+0.019}} & \textcolor{blue}{\textbf{+0.005}} & \textcolor{blue}{\textbf{+0.008}}\\
  \hline
  Resnet50 (100$\%$) & 0.838 & 0.868 & 0.916 & 0.931 \\
  w/ JNDMix  & \textbf{0.851}  & \textbf{0.872} & \textbf{0.920} & \textbf{0.932} \\
      & \textcolor{blue}{\textbf{+0.013}} & \textcolor{blue}{\textbf{+0.004}} & \textcolor{blue}{\textbf{+0.004}} & \textcolor{blue}{\textbf{+0.001}}\\
  \hline
  ViT (100$\%$) & 0.776 & 0.816 & 0.901 & 0.915 \\
  w/ JNDMix  & \textbf{0.815} & \textbf{0.858} & \textbf{0.906} & \textbf{0.923} \\
    & \textcolor{blue}{\textbf{+0.039}} & \textcolor{blue}{\textbf{+0.042}} & \textcolor{blue}{\textbf{+0.005}} & \textcolor{blue}{\textbf{+0.008}}\\
  \hline

\end{tabular}

\vspace{-2mm}
\label{tab:baseline}
\end{table}

\begin{table}[t]
\footnotesize
\centering
\vspace{-2mm}
\caption{Data-efficiency improvements with JNDMix in the \textbf{state-of-the-art models HyperIQA and DBCNN}.}
\vspace{-2mm}
  \begin{tabular}{lrrrr}
  \hline\specialrule{0em}{1pt}{1pt}
  \multicolumn{1}{c}{\multirow{2}{*}{Method}} & \multicolumn{2}{c}{LIVEC} & \multicolumn{2}{c}{KonIQ-10k} \\ \cline{2-5}\specialrule{0em}{1pt}{1pt}
  \multicolumn{1}{c}{} & SRCC & PLCC & SRCC & PLCC \\ 
  \hline
  HyperIQA (100$\%$) & 0.859 & 0.882 & 0.906 & 0.917 \\
  w/ JNDMix  & \textbf{0.870}  & \textbf{0.883} & \textbf{0.911} & \textbf{0.924} \\
      & \textcolor{blue}{\textbf{+0.011}} & \textcolor{blue}{\textbf{+0.001}} & \textcolor{blue}{\textbf{+0.005}} & \textcolor{blue}{\textbf{+0.007}}\\
  \hline
  HyperIQA (50$\%$) & 0.820 & 0.843 & 0.892 & 0.908 \\
  w/ JNDMix  & \textbf{0.821}  & \textbf{0.855} & \textbf{0.902} & \textbf{0.913} \\
      & \textcolor{blue}{\textbf{+0.001}} & \textcolor{blue}{\textbf{+0.012}} & \textcolor{blue}{\textbf{+0.010}} & \textcolor{blue}{\textbf{+0.005}}\\
  \hline
  HyperIQA (25$\%$) & 0.787 & 0.809 & 0.880 & 0.896 \\
  w/ JNDMix  & \textbf{0.790} & \textbf{0.818} & \textbf{0.885} & \textbf{0.900} \\
      & \textcolor{blue}{\textbf{+0.003}} & \textcolor{blue}{\textbf{+0.009}} & \textcolor{blue}{\textbf{+0.005}} & \textcolor{blue}{\textbf{+0.004}}\\
  \hline
  HyperIQA (10$\%$) & 0.700 & 0.736 & 0.851 & 0.866 \\
  w/ JNDMix  & \textbf{0.740}  & \textbf{0.773} & \textbf{0.861} & \textbf{0.874} \\
      & \textcolor{blue}{\textbf{+0.040}} & \textcolor{blue}{\textbf{+0.037}} & \textcolor{blue}{\textbf{+0.010}} & \textcolor{blue}{\textbf{+0.008}}\\
  \hline
  DBCNN (100$\%$) & 0.851 & 0.869 & 0.906 & 0.923 \\
  w/ JNDMix  & \textbf{0.854}  & \textbf{0.876} & \textbf{0.915} & \textbf{0.930} \\
    & \textcolor{blue}{\textbf{+0.003}} & \textcolor{blue}{\textbf{+0.007}} & \textcolor{blue}{\textbf{+0.009}} & \textcolor{blue}{\textbf{+0.007}}\\
  \hline
  DBCNN (50$\%$) & 0.814 & 0.843 & 0.902 & 0.918 \\
  w/ JNDMix  & \textbf{0.815}  & \textbf{0.849} & \textbf{0.907} & \textbf{0.924} \\
    & \textcolor{blue}{\textbf{+0.001}} & \textcolor{blue}{\textbf{+0.006}} & \textcolor{blue}{\textbf{+0.005}} & \textcolor{blue}{\textbf{+0.006}}\\
  \hline
  DBCNN (25$\%$) & 0.764 & 0.793 & 0.888 & 0.907 \\
  w/ JNDMix  & \textbf{0.787} & \textbf{0.812} & \textbf{0.893} & \textbf{0.910} \\
      & \textcolor{blue}{\textbf{+0.023}} & \textcolor{blue}{\textbf{+0.019}} & \textcolor{blue}{\textbf{+0.005}} & \textcolor{blue}{\textbf{+0.003}}\\
  \hline
  DBCNN (10$\%$) & 0.699 & 0.740 & 0.869 & 0.889 \\
  w/ JNDMix  & \textbf{0.718}  & \textbf{0.756} & \textbf{0.873} & \textbf{0.893} \\
      & \textcolor{blue}{\textbf{+0.019}} & \textcolor{blue}{\textbf{+0.016}} & \textcolor{blue}{\textbf{+0.004}} & \textcolor{blue}{\textbf{+0.004}}\\
  \hline
\end{tabular}

\vspace{-1mm}
\label{tab:hydb}
\end{table}

\subsection{Experimental Settings}
\label{ssec:setting}
\noindent\textbf{Datasets and Evaluation Metrics.}
We assess different NR-IQA models and JNDMix on two commonly used IQA datasets: LIVEC \cite{ghadiyaram2015livec} and KonIQ-10k \cite{hosu2020koniq}. LIVEC contains 1162 images taken in the real world by various photos using various camera devices, resulting in complex and composite distortions. KonIQ-10k consists of 10073 images chosen from the large public multimedia dataset YFCC100m\cite{thomee2016yfcc100m}.

Two commonly used evaluation metrics Spearman’s rank order correlation coefficient (SRCC) and Pearson’s linear correlation coefficient (PLCC) are adopted to evaluate the performance of NR-IQA models. 

These datasets are randomly split into 80\% images for training, and 20\% images for testing. In data-efficient experiments, we always use the 20\% images for testing regardless of different fractions of the training dataset. We conduct 10 times of this random dataset splitting operation for each experiment and the average SRCC and PLCC values are reported to evaluate performance.
\vspace{3pt}

\noindent\textbf{Models and Implementation Details.}
To evaluate our proposed data augmentation method, we utilize JNDMix on these three representative state-of-the-art NR-IQA models: DBCNN \cite{zhang2018blind}, HyperIQA \cite{su2020blindly}, MANIQA \cite{yang2022maniqa}. Also, three baseline models are adopted to evaluate JNDMix effectiveness: Resnet34, Resnet50 \cite{he2016res} and Vision Transformer (ViT) \cite{dosovitskiy2020vit}, which are pretrained on ImageNet \cite{deng2009imagenet}. The JND images are generated by RGB-JND \cite{jin2022full} model to generate JND noise. For rigorous comparison, the training settings for all models trained with JNDMix are the same as without it, including learning rate schedule, weight decay, data pre-processing and augmentations, etc. For three state-of-the-art models, we utilize their original codebases and training settings to train. For Resnet34 and Resnet50, we utilize AdamW \cite{loshchilov2018adamw} optimization with a mini-batch of 24, and set the base learning rate to 5e-5 with a cosine decay schedule. For ViT, we utilize AdamW optimization with a mini-batch of 8, and set base learning rate to 1e-5 with a cosine decay schedule. 

\subsection{JNDMix improves IQA benchmarks state-of-the-art}
\label{ssec:m-p}
We train three representative state-of-the-art models in NR-IQA task, DBCNN, HyperIQA, and MANIQA, to study if JNDMix can improve state-of-the-art NR-IQA models on two IQA benchmarks, LIVEC and KonIQ-10k. Table~\ref{tab:sota} shows the results of applying JNDMix on the state-of-the-art models. All models using JNDMix have significant improvements based on their original state-of-the-art performance. Specifically, it provides a gain of +0.011 SRCC and +0.01 PLCC on LIVEC, resulting in MANIQA with the state-of-the-art performance of 0.897 SRCC and 0.918 PLCC on LIVEC, as well as achieve the state-of-the-art performance of 0.927 SRCC and 0.943 PLCC on KonIQ-10k.
\par
Additionally, we reveal that JNDMix improves the performance of baseline models such as Resnet34, Resnet50 and Vision Transformer (ViT). Table~\ref{tab:baseline} illustrates that we get excellent improvements over these baseline models. JNDMix improves ViT from 0.816 to 0.858 with a gain of +0.042 in PLCC on LIVEC, and also achieves a great improvement on Resnet34 with a gain of +0.019 PLCC.

\begin{table}[t]
\footnotesize
\centering
\vspace{-2mm}
\caption{Data-efficiency improvements with JNDMix in the \textbf{baseline models Resnet34 and Resnet50}.}
\vspace{-2mm}
\begin{tabular}{lrrrr}
  \hline\specialrule{0em}{1pt}{1pt}
  \multicolumn{1}{c}{\multirow{2}{*}{Method}} & \multicolumn{2}{c}{LIVEC} & \multicolumn{2}{c}{KonIQ-10k} \\ \cline{2-5}\specialrule{0em}{1pt}{1pt}
  \multicolumn{1}{c}{} & SRCC & PLCC & SRCC & PLCC \\ 
  \hline
  Resnet34 (100$\%$) & 0.824 & 0.849 & 0.905 & 0.920 \\
  w/ JNDMix  & \textbf{0.834}  & \textbf{0.855} & \textbf{0.910} & \textbf{0.928} \\
      & \textcolor{blue}{\textbf{+0.010}} & \textcolor{blue}{\textbf{+0.006}} & \textcolor{blue}{\textbf{+0.005}} & \textcolor{blue}{\textbf{+0.008}}\\
  \hline
  Resnet34 (50$\%$) & 0.788 & 0.816 & 0.890 & 0.908 \\
  w/ JNDMix  & \textbf{0.811}  & \textbf{0.834} & \textbf{0.898} & \textbf{0.914} \\
      & \textcolor{blue}{\textbf{+0.023}} & \textcolor{blue}{\textbf{+0.018}} & \textcolor{blue}{\textbf{+0.008}} & \textcolor{blue}{\textbf{+0.006}}\\
  \hline
  Resnet34 (25$\%$) & 0.738 & 0.746 & 0.869 & 0.888 \\
  w/ JNDMix  & \textbf{0.758} & \textbf{0.766} & \textbf{0.881} & \textbf{0.900} \\
      & \textcolor{blue}{\textbf{+0.020}} & \textcolor{blue}{\textbf{+0.020}} & \textcolor{blue}{\textbf{+0.012}} & \textcolor{blue}{\textbf{+0.012}}\\
  \hline
  Resnet34 (10$\%$) & 0.635 & 0.609 & 0.850 & 0.871 \\
  w/ JNDMix  & \textbf{0.674}  & \textbf{0.653} & \textbf{0.859} & \textbf{0.878} \\
      & \textcolor{blue}{\textbf{+0.039}} & \textcolor{blue}{\textbf{+0.044}} & \textcolor{blue}{\textbf{+0.009}} & \textcolor{blue}{\textbf{+0.007}}\\
  \hline
   Resnet50 (100$\%$) & 0.835 & 0.868 & 0.916 & 0.931 \\
  w/ JNDMix  & \textbf{0.842}  & \textbf{0.872} & \textbf{0.920} & \textbf{0.932} \\
      & \textcolor{blue}{\textbf{+0.007}} & \textcolor{blue}{\textbf{+0.004}} & \textcolor{blue}{\textbf{+0.004}} & \textcolor{blue}{\textbf{+0.001}}\\
  \hline
  Resnet50 (50$\%$) & 0.801 & 0.827 & 0.902 & 0.917 \\
  w/ JNDMix  & \textbf{0.812}  & \textbf{0.840} & \textbf{0.910} & \textbf{0.923} \\
      & \textcolor{blue}{\textbf{+0.011}} & \textcolor{blue}{\textbf{+0.013}} & \textcolor{blue}{\textbf{+0.008}} & \textcolor{blue}{\textbf{+0.006}}\\
  \hline
  Resnet50 (25$\%$) & 0.751 & 0.768 & 0.886 & 0.902 \\
  w/ JNDMix  & \textbf{0.767} & \textbf{0.794} & \textbf{0.892} & \textbf{0.908} \\
      & \textcolor{blue}{\textbf{+0.016}} & \textcolor{blue}{\textbf{+0.026}} & \textcolor{blue}{\textbf{+0.006}} & \textcolor{blue}{\textbf{+0.006}}\\
  \hline
  Resnet50 (10$\%$) & 0.657 & 0.682 & 0.863 & 0.880 \\
  w/ JNDMix  & \textbf{0.681}  & \textbf{0.695} & \textbf{0.868} & \textbf{0.884} \\
      & \textcolor{blue}{\textbf{+0.024}} & \textcolor{blue}{\textbf{+0.013}} & \textcolor{blue}{\textbf{+0.005}} & \textcolor{blue}{\textbf{+0.004}}\\
  \hline
\end{tabular}
\vspace{-2mm}
\label{tab:resnet}
\end{table}
\subsection{JNDMix helps data-efficiency}
\label{ssec:d-e}
In this section, we demonstrate that JNDMix is effective to improve model performance across different dataset sizes and helps data efficiency of models. Table~\ref{tab:hydb} illustrates that JNDMix always improves the perfomance of HyperIQA and DBCNN across all fractions of LIVEC and KonIQ-10k, which implies the data efficiency improvement of the state-of-the-art models. JNDMix provides a surprising improvement of +0.04 SRCC in the low data regime (10$\%$ of data) while still being effective with a improvement of +0.007 PLCC in the high data regime on LIVEC. We detect that the performance growth brought by JNDMix is increasing with the decrease of the dataset sizes, which is also reflected in the difference in performance improvements between the two datasets, where the images of LIVEC is 10 times less than those of KonIQ-10k. Table~\ref{tab:resnet} reveals that JNDMix also greatly helps data efficiency of baseline models with the maximum gain of +0.044 PLCC in the low data regime (10$\%$ of data).

\subsection{JNDMix improves generalization ability}
To show the generaliazation ability of NR-IQA models, the cross dataset evaluations are performed. We train MANIQA on different fractions of KonIQ-10k and test it on LIVEC. Table~\ref{tab:cross} proves that JNDMix is effective to improve generalization ability of NR-IQA models across different dataset sizes.

\label{ssec:g-a}

\begin{table}[t]
\footnotesize
\centering
\vspace{-2mm}
\caption{Results of cross dataset evaluations trained on different fractions of datasets adopting the state-of-the-art model MANIQA.}
\vspace{-1mm}
\begin{tabular}{cclrr}
\hline\specialrule{0em}{1pt}{1pt}
Training                    & Testing                                         & \multicolumn{1}{c}{Method} & SRCC & PLCC \\ \specialrule{0em}{1pt}{1pt}\hline
\multirow{12}{*}{KonIQ-10k} & \multicolumn{1}{c}{\multirow{12}{*}{LIVEC}}    & MANIQA (100\%)             &    0.844  & 0.868      \\
                            &                           & w/  JNDMix                 & \textbf{0.856} & \textbf{0.880}      \\
                            & \multicolumn{1}{c}{}                           &                      & \textcolor{blue}{\textbf{+0.012}} &\textcolor{blue}{\textbf{+0.012}}      \\ \cline{3-5} 
                            & \multicolumn{1}{c}{}                           & MANIQA (50\%)              & 0.837  & 0.858 \\
                            & \multicolumn{1}{c}{}                           & w/  JNDMix                 & \textbf{0.843} & \textbf{0.863} \\
                            & \multicolumn{1}{c}{}                           &                            & \textcolor{blue}{\textbf{+0.006}} &\textcolor{blue}{\textbf{+0.005}}      \\ \cline{3-5} 
                            & \multicolumn{1}{c}{}                           & MANIQA (25\%)              & 0.828 & 0.854 \\
                            & \multicolumn{1}{c}{}                           & w/ JNDMix                  & \textbf{0.837} & \textbf{0.862} \\
                            & \multicolumn{1}{c}{}                           &                            & \textcolor{blue}{\textbf{+0.009}} &\textcolor{blue}{\textbf{+0.008}}      \\ \cline{3-5} 
                            & \multicolumn{1}{c}{}                           & MANIQA (10\%)              & 0.797 & 0.828 \\
                            & \multicolumn{1}{c}{}                           & w/  JNDMix                 & \textbf{0.808} & \textbf{0.836} \\
                            & \multicolumn{1}{c}{}                           &                            & \textcolor{blue}{\textbf{+0.011}} &\textcolor{blue}{\textbf{+0.008}}      \\ \hline
\multirow{3}{*}{LIVEC}      & \multicolumn{1}{c}{\multirow{3}{*}{KonIQ-10k}} & MANIQA (100\%)             & 0.790  &  0.849  \\
                            & \multicolumn{1}{c}{}                           & w/ JNDMix                  & \textbf{0.803} & \textbf{0.861} \\
                            & \multicolumn{1}{c}{}                           &                            & \textcolor{blue}{\textbf{+0.013}}    &\textcolor{blue}{\textbf{+0.012}}      \\ \hline
\end{tabular}
\label{tab:cross}
\end{table}

\begin{table}[t]
\small
\centering

\caption{Performance of MANIQA with different noise injection on 100\% LIVEC.}
\vspace{-1mm}
\begin{tabular}{lrr}
\hline
\specialrule{0em}{1pt}{1pt}
\multicolumn{1}{c}{Method} & SRCC  & PLCC  \\ \specialrule{0em}{1pt}{1pt}\hline
MANIQA                                                                                     & 0.886 & 0.908 \\
w/ JNDMix                                                                                    & \textbf{0.897} & \textbf{0.918} \\ \hline
w/ JND injection                                                                             & 0.891 & 0.911 \\
w/ Gaussian noise injection                                                                  & 0.796 & 0.825 \\ \hline
\end{tabular}
\vspace{-3mm}
\label{tab:ablation}
\end{table}
\subsection{Ablation Study}
To study the contribution of JND noise injection, we conduct ablation study by injecting different noises. Table~\ref{tab:ablation} shows the performance of MANIQA with different noise injection trained on 100\% LIVEC. `JND injection' means we do not use JND noise generated by multiplying the JND image with a resampled random ratio, but directly add the JND image to the training image. `Gaussian noise injection' means we inject the Gaussian noise to generate the new training image. The results reveal that injecting Gaussian noise leads to performance degradation and adding the JND image results in smaller improvement than JNDMix.

\section{Conclusion}
Data augmentation is vital to many vision tasks, and however, no effective data augmentation method specifically designed for NR-IQA task is proposed. In this paper, we are the first to propose a general data augmentation method for NR-IQA task, and rigorously study the JNDMix, finding it highly effective and general. 
JNDMix works well across a variety of different model architectures and helps their data efficiency, both on LIVEC and KonIQ-10k. 
The JNDMix augmentation method is simple and can be flexibly plugged into any NR-IQA model codebase. We wish that our method will be widely used in NR-IQA task and the lack of data augmentation methods in NR-IQA task will receive attention.
\vfill\pagebreak

\bibliographystyle{IEEEbib}
\small{
\bibliography{strings,refs}
}

\end{document}